\documentclass[runningheads]{llncs}
\usepackage{ifxetex}
\ifxetex\else
\usepackage[utf8]{inputenc}
\fi
\usepackage[T1]{fontenc}
\usepackage{microtype}
\usepackage{graphicx}
\usepackage{color}
\usepackage{xcolor}
\usepackage[colorlinks,allcolors=black,pdfencoding=auto,psdextra]{hyperref}

\usepackage[inline,shortlabels]{enumitem}
\usepackage{lipsum}
\usepackage{mdframed}
\mdfsetup{skipabove=0pt,skipbelow=0pt}
\usepackage{wrapfig}

\usepackage{amsmath}
\usepackage{mathtools}
\usepackage{eqnarray}

\ifxetex%
\usepackage{fontspec}
\usepackage{unicode-math}
\defaultfontfeatures{Mapping=tex-text,Scale=MatchLowercase,Ligatures=TeX}
\else
\DeclareUnicodeCharacter{0308}{\'{u}}
\fi

\usepackage{tikz}
\usetikzlibrary{%
  arrows,
  arrows.meta,
  calc,
  chains,
  fit,
  math,
  matrix,
  patterns,
  positioning,
  quotes,
  shapes.arrows,
  shapes.geometric,
  shapes.misc,
  trees,
}
\tikzset{%
  font=\sffamily,
  x=1em,
  y=1em,
  >=stealth',
  node distance=2em,
}
\tikzstyle{n}=[% node
  draw,
  rounded rectangle,
  rounded rectangle arc length=120,
  font=\ttfamily\footnotesize,
  inner sep=0pt,
  outer sep=0pt]
\tikzstyle{l}=[% literal
  font=\ttfamily\footnotesize,
  inner sep=0pt,
  outer sep=0pt]
\tikzstyle{e}=[% edge
  fill=white,
  font=\ttfamily\scriptsize,
  inner sep=0pt,
  outer sep=1pt]

\usepackage[frozencache,cachedir=_minted]{minted}
\usepackage{minted}
\usepackage{etoolbox}
\makeatletter
\patchcmd{\minted@colorbg}{\medskip}{\smallskip}{}{}
\patchcmd{\endminted@colorbg}{\medskip}{\smallskip}{}{}
\makeatother
\definecolor{bg}{rgb}{0.95,0.95,0.95}
\definecolor{rl}{rgb}{0.5,0.5,0.5}
\newmintinline[code]{text}{escapeinside=!!}
\newminted{text}{%
  python3=true,
  fontsize=\footnotesize,
  tabsize=2,
  bgcolor=bg,
  breaklines,
  linenos=false,
  numbersep=3pt,
  escapeinside=!!}
\newmintinline[py]{python}{escapeinside=!!}
\newminted{pycon}{%
  python3=true,
  fontsize=\footnotesize,
  tabsize=2,
  bgcolor=bg,
  breaklines,
  linenos=false,
  numbersep=3pt,
  escapeinside=!!}
\newmintedfile{pycon}{%
  python3=true,
  fontsize=\footnotesize,
  tabsize=2,
  bgcolor=bg,
  breaklines,
  linenos=false,
  numbersep=3pt,
  escapeinside=!!}

\newcommand*\KIFURL{\url{https://github.com/IBM/kif}}

\begin{document}

\title{KIF\@: A Wikidata-Based Framework for
  Integrating Heterogeneous Knowledge Sources}
\titlerunning{Wikidata-Based Framework for Integrating Knowledge Sources}

\author{
  Guilherme Lima
  \and
  João M.~B.~Rodrigues
  \and
  Marcelo Machado
  \and
  Elton Soares
  \and\\
  Sandro R.~Fiorini
  \and
  Raphael Thiago
  \and
  Leonardo G.~Azevedo
  \and\\
  Viviane T.~da Silva
  \and
  Renato Cerqueira}

\authorrunning{G.~Lima et al.}

\institute{
  IBM Research Brazil, Rio de Janeiro, Brazil\\
  \email{\{guilherme.lima,joao.bessa,mmachado,eltons,srfiorini\}@ibm.com},\
  \email{\{raphaelt,lga,vivianet,rcerq\}@br.ibm.com}}

\maketitle

\begin{abstract}
  We present a Wikidata-based framework, called KIF, for virtually integrating heterogeneous knowledge sources.
  KIF is written in Python and is released as open-source.
  It leverages Wikidata's data model and vocabulary plus user-defined mappings to construct a unified view of the underlying sources while keeping track of the context and provenance of their statements.
  The underlying sources can be triplestores, relational databases, CSV files, etc., which may or may not use the vocabulary and RDF encoding of Wikidata.
  The end result is a virtual knowledge base which behaves like an ``extended Wikidata'' and which can be queried using a simple but expressive pattern language, defined in terms of Wikidata's data model.
  In this paper, we present the design and implementation of KIF, discuss how we have used it to solve a real integration problem in the domain of chemistry (involving Wikidata, PubChem, and IBM CIRCA), and present experimental results on the performance and overhead of KIF\@.

\end{abstract}

\section{Introduction}%
\label{sec:introduction}

Knowledge source integration is the problem of meaningfully combining multiple knowledge sources.
The problem is harder when (i) the sources are heterogeneous, i.e., adopt different vocabularies, formats, storage technologies, etc.; and (ii) the intended integration is virtual, i.e., is to be done dynamically, at query time.
In the Semantic Web, where ``knowledge source'' usually means a set of OWL ontologies, the integration problem is often reduced to the ontology matching problem~\cite{Osman-I-2021}.
In practice, however, determining correspondences between concepts and properties in ontologies is just one of the many issues involved.
More often than not, the sources to be integrated don't support OWL or even RDF (think of graph databases, relational databases, REST APIs, CSV dumps, etc.) and either are just too large or change too frequently to be converted and ingested statically into a single knowledge base.

A scenario like this requires a more comprehensive solution.
One that combines virtualization mechanisms to provide federated access to the sources, mapping mechanisms to reconcile differences in their vocabularies, and provenance mechanisms to enable telling which piece of knowledge came from which source.
Although there are in the literature proposals to tackle each of these tasks separately, few attempt to provide an integrated solution.

In this paper, we present KIF\footnote{\KIFURL}, an open-source Python framework that uses Wikidata's data model and vocabulary~\cite{Vrandecic-D-2014} as a lingua franca to integrate heterogeneous knowledge sources.
KIF is our attempt at a comprehensive solution to the hard version of the knowledge integration problem---the version in which the sources are heterogeneous and the integration is virtual.

The core abstraction of KIF is the store.
A \emph{store} is an interface to a ``Wikidata view'' of a knowledge source, obtained by interpreting the source's content as a set of Wikidata-like statements.
For example, KIF's built-in CSV store constructs Wikidata view of a CSV file by interpreting each of its cells (line-column pair) as a statement where the subject is given by the entity described by the line, the property is given by the header of the column, and the value by the content of the cell.
Another built-in store type is the SPARQL store.
It constructs a Wikidata view of a SPARQL endpoint by interpreting certain patterns in the endpoint's RDF graph as Wikidata-like statements.

% Another built-in store type is the SPARQL store which constructs a Wikidata view of a given SPARQL endpoint, which may or may not speak the Wikidata dialect of RDF.

% In the latter case, the user must provide a \emph{mapping} indicating how certain patterns in the endpoint's RDF are to be interpreted as Wikidata-like statements.

Having Wikidata as the target brings some advantages.
First, KIF inherits a tried-and-tested data model with native support for structured data values plus the notions of references and qualifiers, used to represent provenance and context information.
Second, if desired, one can reuse Wikidata's vast vocabulary, which at the time of writing has more than 110M items and 11K properties, covering various domains.
Third, one can easily combine statements produced by any KIF store with statements coming from Wikidata itself, which can be accessed via an ordinary SPARQL store.
Such a combination is done using a \emph{mixer} store, which virtually merges statements of one or more child stores.

The key to the flexibility of the store abstraction lies in its query interface.
KIF stores are queried using a simple but expressive \emph{pattern language} defined in terms of Wikidata's data model.
KIF includes a pattern compiler targeting SPARQL which can be parameterized with custom mappings to generate queries in RDF encodings other than Wikidata's.
A mapping to the RDF encoding of PubChem~\cite{Kim-S-2023} (a large public base of chemical knowledge) is also included in the distribution.
As we discuss later, this mapping was used together with other things to build an application that integrates statements about chemical compounds coming from PubChem, Wikidata, and IBM CIRCA~\cite{IBM-CIRCA} (a relational database of chemical data extracted from patents and other sources).

The rest of the paper is organized as follows.
Section~\ref{sec:background} presents some background on Wikidata.
Section~\ref{sec:design-and-implementation} presents the design and implementation of KIF\@.
Section~\ref{sec:use-case-and-evaluation} discusses the evaluation of KIF over a use case in the domain of chemistry.
Section~\ref{sec:related-work} discusses some related work.
Section~\ref{sec:conclusion} presents our conclusions and future work.

% LocalWords:  CSV KIF lingua franca PubChem KIF's


%%% Local Variables:
%%% mode: latex
%%% TeX-engine: xetex
%%% TeX-master: "main"
%%% eval: (visual-line-mode 1)
%%% End:

\section{Background}%
\label{sec:background}

Wikidata~\cite{Vrandecic-D-2014} is a sister project of Wikipedia and it's also one of largest bases of structured knowledge on the Web.
Although we have been using the term ``Wikidata data model'', the data model used by Wikidata actually comes from Wikibase, which is the open-source framework underlying Wikidata.
Wikibase is a project of its own.
It can be used to create other knowledge bases following the same data model as Wikidata but with different content and purposes~\cite{Diefenbach-D-2021}.

\subsection{Wikibase Data Model}%
\label{sec:background:data-model}

\begin{figure}[ht]
  \centering
  \includegraphics[width=.95\textwidth]{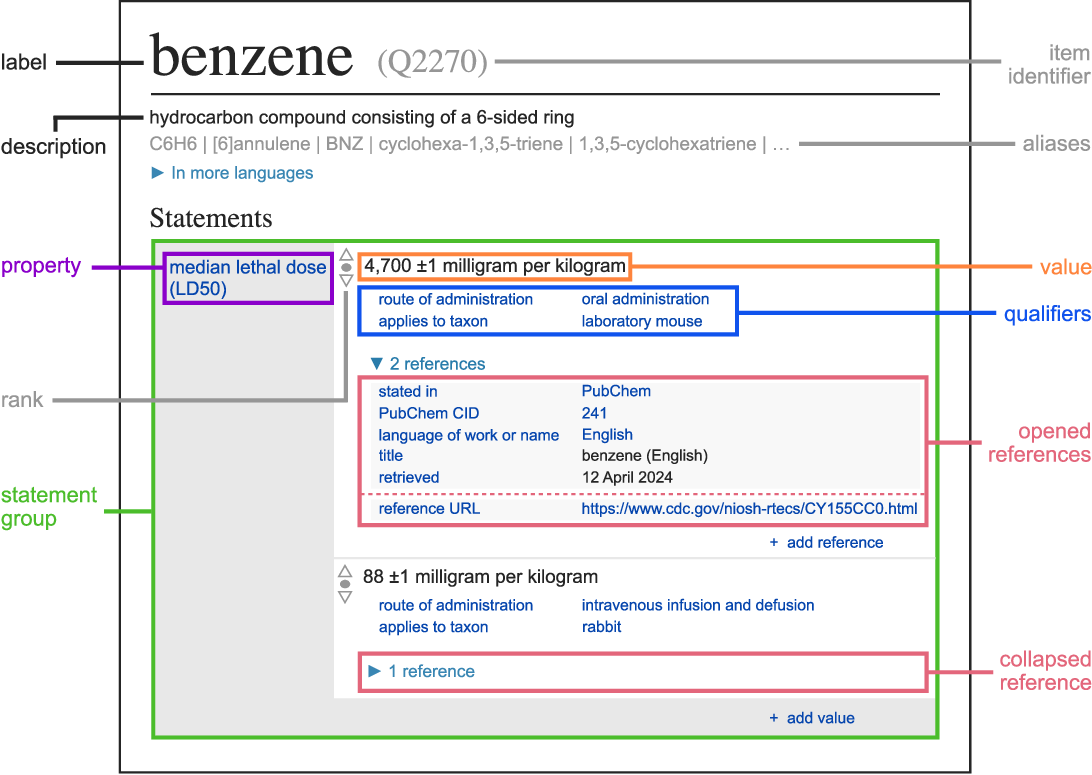}%
  \caption{Part of Wikidata's entity page of benzene. (Adapted from~\cite{Odell-J-2022}.)}%
  \label{fig:benzene}
  \vskip-.75\baselineskip%
\end{figure}

Wikibase's data model~\cite{Wikibase-DataModel}
%%
%\footnote{\url{https://www.mediawiki.org/wiki/Wikibase/DataModel}}
%%
consists of entities and statements about entities.
In Wikidata's UI, statements are grouped into ``entity pages''.
Figure~\ref{fig:benzene} shows the page of entity Q2270, which stands for the chemical compound benzene.\footnote{\url{https://www.wikidata.org/wiki/Q2270}}
Every entity has a label, a description, and one or more aliases.
In Figure~\ref{fig:benzene}, these are shown in the header at the top of the page.

After the header comes the ``Statements'' section which groups statements about the entity being described.
A \emph{statement} consists of two parts: subject and snak.
The \emph{subject} is the entity about which the statement is made.
The \emph{snak} is the statement's claim.
It associates a property with either a specific value, some unspecified value, or no value.

Figure~\ref{fig:benzene} depicts two statements which can be read as follows:
\begin{align}
  \label{eq:stmt1}
  &\text{``benzene has an LD50 of 4{,}699--4{,}701 milligrams per kilogram''}\\
  \label{eq:stmt2}
  &\text{``benzene has an LD50 of 87--89 milligrams per kilogram''}
\end{align}
LD50 (or median lethal dose) is a toxicity unit that measures the dose of a substance that is required to kill half the members of a tested population.

The subject of statements~\eqref{eq:stmt1} and~\eqref{eq:stmt2} is the same, ``benzene'' (Q2270).
Their snak is of the form property-value.
The property of both is ``median lethal dose (LD50)'' (P2240).
The value of~\eqref{eq:stmt1} is ``4{,}700 $\pm$1 mg/kg'' and the value of~\eqref{eq:stmt2} is ``88 $\pm$1 mg/kg''.
Note that the data model distinguishes between items (identified with ``Q'') and properties (identified with ``P'').
Only the latter can occur as the first component of snaks.

In Python, using the KIF constructors (see Figure~\ref{fig:grammar}), statements~\eqref{eq:stmt1} and~\eqref{eq:stmt2} are written as follows:
\newcommand*\X[1]{$\langle\text{\scriptsize{\textrm{#1}}}\rangle$}%
\begin{pyconcode}
>>> stmt1 = Statement(Item(!\X{benzene}!),
...    ValueSnak(Property(!\X{LD50}!), Quantity(4700, !\X{mg/kg}!, 4699, 4701)))

>>> stmt2 = Statement(Item(!\X{benzene}!),
...    ValueSnak(Property(!\X{LD50}!), Quantity(88, !\X{mg/kg}!, 87, 89)))
\end{pyconcode}
We write $\langle{x}\rangle$ for the URL of entity $x$, e.g., $\langle\text{benzene}\rangle$ stands for \url{http://www.wikidata.org/entity/Q2270}.

Back to Figure~\ref{fig:benzene}, the qualifiers and references associated with each statement are shown below the statement's value (see the boxes ``qualifiers'' and ``opened references'').
\emph{Qualifiers} are extra snaks which qualify what is being said by the statement's main snak; \emph{references} (or \emph{reference records}) are sets of snaks which keep provenance information.

The qualifiers of statement~\eqref{eq:stmt1} shown in Figure~\ref{fig:benzene} are written as follows:
\begin{pyconcode}
>>> qualifiers_of_stmt1 = [
...    ValueSnak(Property(!\X{route of administration}!), Item(!\X{oral administration}!))
...    ValueSnak(Property(!\X{applies to taxon}!), Item(!\X{laboratory mouse}!))]
\end{pyconcode}
Note that the qualifiers in this case convey information that is crucial to interpret the statement, namely, the route of administration and the affected animal.

The references shown for statement~\eqref{eq:stmt1} in Figure~\ref{fig:benzene} are written as follows:
\begin{pyconcode}
>>> references_of_stmt1 = [
... ReferenceRecord(    # 1st reference
...    ValueSnak(Property(!\X{stated in}!), Item(!\X{PubChem}!)),
...    ValueSnak(Property(!\X{PubChem CID}!), String('241')),
...    ValueSnak(Property(!\X{language of work or name}!), Item(!\X{English}!)),
...    ValueSnak(Property(!\X{retrieved}!), Time('2024-04-12'))),
... ReferenceRecord(    # 2nd reference
...    ValueSnak(Property(!\X{reference URL}!), IRI('http://www.cdc.gov...')))]
\end{pyconcode}

\begin{figure}[ht]
  \newcommand*\xcode[1]{\text{\code{#1}}}
  \newcommand*\NT[1]{\ensuremath{\text{\emph{#1}}}}%
  \newcommand*\EQ{\mathord{::=}}%
  \begin{mdframed}%
  \abovedisplayskip=-.5em
  \belowdisplayskip=-.5em
  \def\arraystretch{1.25}
  \begin{equationarray*}{rll}%
    \NT{stmt}      \ \EQ &\ \xcode{Statement(!\NT{entity}!, !\NT{snak}!)}
                         &\text{--- claim about \NT{entity}}\\
    \NT{entity}    \ \EQ &\ \NT{item} \mid \NT{property}\\
    \NT{item}      \ \EQ &\ \xcode{Item(!\NT{iri}!)}
                         &\text{--- person or thing}\\
    \NT{property}  \ \EQ &\ \xcode{Property(!\NT{iri}!)}
                         &\text{--- (binary) relation}\\
    \NT{snak}      \ \EQ &\ \xcode{ValueSnak(!\NT{property}!, !\NT{value}!)}
                         &\text{--- ``\NT{property} has \NT{value}''}\\
                   \ \mid&\ \xcode{SomeValueSnak(!\NT{property}!)}
                         &\text{--- ``\NT{property} has some value''}\\
                   \ \mid&\ \xcode{NoValueSnak(!\NT{property}!)}
                         &\text{--- ``\NT{property} has no value''}\\
    %                      %%
    \NT{value}     \ \EQ &\ \NT{entity} \mid \NT{data-value}\\
    \NT{data-value}\ \EQ &\multicolumn{2}{l}{%
      \ \NT{iri} \mid \NT{text} \mid \NT{string} \mid
      \NT{external-id} \mid \NT{quantity} \mid \NT{time}}\\
    \NT{iri}       \ \EQ &\ \xcode{IRI(!{$s$}!)}
                         &\text{--- IRI}\\
    \NT{text}      \ \EQ &\ \xcode{Text(!{$s$}!, !\NT{lang}${?}$!)}
                         &\text{--- text in a given language}\\
    \NT{string}    \ \EQ &\ \xcode{String(!{$s$}!)}
                         &\text{--- string}\\
    \NT{external-id}\ \EQ&\ \xcode{ExternalId(!{$s$}!)}
                         &\text{--- external id}\\
    %                      %%
    \NT{quantity}  \ \EQ &\
                   \xcode{Quantity(!{$n$}!, !\NT{item}$?$!, !{$n?$}!, !{$n?$}!)}
                         &\text{--- numerical quantity}\\
    \NT{time}      \ \EQ &\
                   \xcode{Time(!{$ts$}!, !{$i?$}!, !{$i?$}!, !\NT{item}$?$!)}
                         &\text{--- date or time}\\
    \NT{reference} \ \EQ&\ \xcode{ReferenceRecord(!\NT{snak}$+$!)}\\
    \NT{rank}      \ \EQ&\ \xcode{Preferred} \mid \xcode{Normal} \mid
                             \xcode{Deprecated}
  \end{equationarray*}%
  \end{mdframed}%
  \caption{Constructors of data model objects in KIF\@.  ``${?}$'' means zero-or-one; ``$+$'' means one-or-more; $s$ is a Python string; $\NT{lang}$ is a Python string containing language tag such as ``en''; $n$ is a number; $i$ is an integer; and $ts$ is a date-time timestamp.}%
  \label{fig:grammar}%
  %\vskip-1em
\end{figure}


%%% Local Variables:
%%% mode: latex
%%% TeX-engine: xetex
%%% TeX-master: "../main"
%%% eval: (visual-line-mode 1)
%%% End:

The last piece of metadata associated with statements is the \emph{rank} which can be either ``preferred'', ``normal'', or ``deprecated''.
In Figure~\ref{fig:benzene}, the rank is represented symbolically by the two triangles and circle which occur on the left of the statement's value.
A filled upper triangle means preferred rank; a filled circle means normal rank; and a filled lower triangle means deprecated rank.
As can be seen in Figure~\ref{fig:benzene}, statements~\eqref{eq:stmt1} and~\eqref{eq:stmt2} have normal rank.

\subsection{Wikidata RDF Encoding}%
\label{sec:background:rdf}

Wikidata defines an RDF encoding for its data model which is also adopted by Wikibase~\cite{Wikibase-RDF-Dump-Format,Westerinen-A-2024}.
%%
%\footnote{\url{https://www.mediawiki.org/wiki/Wikibase/Indexing/RDF_Dump_Format}}
%%
The format varies slightly depending on whether it is used in a data dump or observed from Wikidata's query service.
The version we describe here is that of the query service.

In Wikidata's RDF encoding, each statement is represented at two levels.
The first level, called \emph{truthy}, keeps a shallow representation of the statement as a single RDF triple.
For example, the truthy encoding of statement~\eqref{eq:stmt1} of Figure~\ref{fig:benzene}, namely, ``benzene (Q2270) has an LD50 (P2240) of 4{,}700$ \pm$1 mg/kg'', consists of the single triple:
\begin{equation}
\label{eq:benzene-truthy}
\begin{tikzpicture}[node distance=6em,baseline]
  \node[n,draw](benzene){\strut\text{wd:Q2270}};
  \node[l,right=of benzene](value){\strut\text{"4700"{\^{}\^{}xsd:decimal}}};
  \draw[->] (benzene) -- node[e,above,inner sep=2pt]{wdt:P2240} (value);
\end{tikzpicture}\tag{$\dagger$}
\end{equation}


%%% Local Variables:
%%% mode: latex
%%% TeX-engine: xetex
%%% TeX-master: "../main"
%%% eval: (visual-line-mode 1)
%%% End:

%%
The namespace \code/wd:/ indicates an entity and \code/wdt:/ indicates a truthy application of a property.
Some statements are fully characterized at the truthy level.
But, as illustrated by~\eqref{eq:benzene-truthy}, this is not always the case.
Note that the unit, lower-, and upper-bounds associated with the value 4700 are not represented in~\eqref{eq:benzene-truthy}.
In general, when the statement's value is a structured data-value, like a quantity or time value, a single literal is used to represent it at the truthy level.
This is the so called \emph{simple value} of the statement.
For quantity values, the simple value is is just the numerical amount.

%%
% TODO: Explain that, i.e., the first argument of the \code/Quantity/ constructor.
%%

The second level of the encoding keeps the full representation of the statement.
It uses reification to capture the \emph{deep value} of the statement plus its qualifiers, references, and rank.
Figure~\ref{fig:benzene-deep} depicts the full representation of statement~\eqref{eq:stmt1} of Figure~\ref{fig:benzene} considering only one qualifier and one reference record.

\begin{figure}[ht]
  \vskip-\baselineskip%
  \centering
  \begin{tikzpicture}[s/.style={n,fill=gray!25},node distance=4em and 4em]
    \node[s](wds){\strut\code/wds:_/};
    \node[n,above left=of wds](benzene){\strut wd:Q2270};
    \node[l,above right=of wds](value){\strut "4700"\rlap{\^{}\^{}xsd:decimal}};
    \draw[->](benzene) to node[e]{wdt:P2240} (value);
    \draw[->](benzene) to[bend right=10] node[e,above]{p:P2240} (wds);
    \begin{scope}
      \node[n,left=of wds,xshift=-2em,text width=6em,align=center]
      (NormalRank){\strut wikibase:\\[-2pt]\strut NormalRank};
      \draw[->](wds) to node[e,text width=3.5em,align=center]
      {wikibase:\\[-2pt]rank} (NormalRank);
    \end{scope}
    \draw[->](wds) to[bend right=10] node[e,above]{ps:P2240} (value);
    \node[s,below right=of wds](wdv){\strut\code/wdv:_/};
    \draw[->](wds) to[bend left=10] node[e,below,yshift=1pt]{psv:P2240} (wdv);
    \draw[->](wdv) to[bend left=10]
    node[e,text width=6em,align=center,yshift=.5em]
    {wikibase:\\[-2pt]quantityAmount} (value);
    \node[n,above right=of wdv,text width=7em,align=center,yshift=1em]
    (QuantityValue){\strut wikibase:\\[-2pt]\strut QuantityValue};
    \draw[->](wdv) to[bend left=10]node[e]{rdf:type} (QuantityValue);
    \node[n,right=of wdv,xshift=3em,yshift=2em](Unit){\strut wd:Q21091747};
    \draw[->](wdv) to[bend left=10]
    node[e,text width=4.7em,align=center,yshift=-.3em]
    {wikibase:\\[-2pt]quantityUnit}(Unit);
    \node[l,below right=of wdv,yshift=2em](LB)
    {\strut"4699"\rlap{\^{}\^{}xsd:decimal}};
    \draw[->](wdv) to[bend right=10] node[e,align=center,xshift=4em,yshift=.4em]
    {wikibase:quantityLowerBound}(LB);
    \node[l,below=of wdv=0em,xshift=-1em](UB)
    {\strut "4701"\rlap{\^{}\^{}xsd:decimal}};
    \draw[->](wdv) to[bend right=10]
    node[e,align=center,text width=7em,xshift=-2em]
    {wikibase:quantity\\[-2pt]UpperBound}(UB);
    \node[n,below=of wds](Oral){\strut wd:Q285166};
    \draw[->](wds) to[bend right=10] node[e]{pq:P636}(Oral);
    \node[s,below left=of wds](wdref){\strut\code/wdref:_/};
    \node[n,below=of wdref](refurl)
    {\strut https://www.cdc.gov/...};
    \draw[->](wds) to[bend right=10] node[e,below,yshift=1pt,xshift=-3em]
    {prov:wasDerivedFrom} (wdref);
    \draw[->](wdref) to[bend right=10] node[e]{pr:P854} (refurl);
  \end{tikzpicture}
  \caption{RDF representation of the statement ``Benzene (Q2270) has an LD50 (P2240) of 4{,}700 $\pm$1 mg/kg (Q21091747)'' considering only the qualifier ``route of administration (P636) is oral administration (Q285166)'' and the reference record ``reference URL (P854) is \url{https://www.cdc.gov/niosh-rtecs/CY155CC0.html}''.}%
  \label{fig:benzene-deep}%
  \vskip-1.5\baselineskip%
\end{figure}


%%% Local Variables:
%%% mode: latex
%%% TeX-engine: xetex
%%% TeX-master: "../main"
%%% eval: (visual-line-mode 1)
%%% End:

In Figure~\ref{fig:benzene-deep}, the shaded nodes are the reified ones.
The single underscore~(\code/_/) indicates that their ids are opaque (hence not shown in the figure).
Node \code/wds:_/ represents the statement.
Predicates \code/p:P2240/ and \code/ps:P2240/ are used to connect the subject ``benzene'' (\code/wd:Q2270/) to the statement and the statement to its simple value, i.e., the number 4700 in decimal notation.

The deep value of the statement is represented by node \code/wdv:_/.
It has type \code/wikibase:QuantityValue/ and is connected to the unit mg/kg (\code/wd:Q21091747/), the lower-bound 4699, and the upper-bound 4701.
The rank of the statement is connected via predicate \code/wikibase:rank/.

Moving to qualifiers, predicate \code/pq:P636/ connects the qualifier ``route of administration'' (P636) with value ``oral administration'' (\code/wd:Q285166/) to the statement.
Finally, predicate \code/prov:wasDerivedFrom/ connects to the statement the reference record represented by node \code/wdref:_/.
Its content (the snak ``reference URL'' (P854) with value ``\code|https://www.cdc.gov/...|'') is encoded using predicate \code/pr:P854/ and a (simple) IRI value.

% LocalWords:  Wikibase Wikibase's UI snak LD snaks KIF IRI eq stmt truthy
% LocalWords:  wd wdt


%%% Local Variables:
%%% mode: latex
%%% TeX-engine: xetex
%%% TeX-master: "main"
%%% eval: (visual-line-mode 1)
%%% End:

\section{Design and Implementation}%
\label{sec:design-and-implementation}

KIF is an integration framework based on Wikidata.
The idea behind it is to use Wikidata to standardize the syntax and possibly the vocabulary of the underlying knowledge sources.
Users can then query the sources through patterns described in terms of Wikidata's data model.
The integration done by KIF is \emph{virtual} in the sense that syntax and vocabulary translations happen dynamically, at query time, guided by user-provided mappings.

As we mentioned before, the core abstraction of KIF is the store.
A \emph{store} is an interface to a knowledge source.
This can be a SPARQL endpoint, REST API, RDF file, CSV file, etc.
The job of the store is to construct a ``Wikidata view'' of the knowledge source.
The prototypical store is the SPARQL store, which we describe next.

\subsection{SPARQL Store}%
\label{sub:sparql-store}

The SPARQL store reads Wikidata-like statements from a given SPARQL endpoint.
%%
%If no mapping is specified at store creation time, it assumes that the endpoint speaks the Wikidata encoding of RDF\@.
%%
Here is how we create a SPARQL store pointing to WDQS, the public SPARQL query service of Wikidata:
\pyconfile[firstline=1,lastline=2,linenos=true]{code/code.pycon}%
At line~1, we import the namespace of KIF, whose Python module is called \code/kif_lib/.
At line~2, we create a new store of type ``sparql'' pointing to WDQS and assign the result to variable \code/Wikidata/.
As we did not pass an explicit mapping to the store constructor, it assumes the endpoint speaks Wikidata's RDF encoding, which is the case for WDQS\@.

We can read statements from the newly created \code/Wikidata/ store as follows:
\pyconfile[firstline=3,lastline=7,linenos=true]{code/code.pycon}%
At lines~3--5, we ask for statements with subject ``benzene'' (Q2270) and property ``median lethal dose (LED50)'' (P2240).
The result is an iterator which is assigned to variable \code/it/.
At line~6, we consume one statement from \code/it/ whose content is shown in line~7.
Note that this is the same statement~\eqref{eq:stmt1} of Section~\ref{sec:background}.

\enlargethispage*{3pt}

\subsection{Patterns}%
\label{sub:patterns}

%%
%The apparent simplicity of the previous example is deceiving.
%%
%Let us take a closer look at what is going on behind the scenes.
%%
As we mentioned, in KIF, queries are specified as patterns defined in terms of Wikidata's data model.
The \code{filter()} call in line~3 above is actually just a wrapper to a \code{match()} call, which evaluates a pattern over the knowledge source.

We can rewrite the previous \code/filter()/ in terms of \code/match()/ as follows:
\pyconfile[firstline=8,lastline=11,linenos=true]{code/code.pycon}%
At line~8, we create the pattern variable \code/x/ and use it to build the pattern \code/pat/ in lines~9--10.
KIF patterns are templates for objects of the data model, i.e., objects in which certain parts are replaced by variables.
Pattern \code/pat/ is a template for statements whose subject is ``benzene'' (Q2270) and whose snak is a value-snak with property ``LD50'' (P2240) and value \code/x/, i.e., any value.

% -- optional -- %
% Although we didn't restrict the type of \code/x/ when we create it in line~8, in KIF, variables that occur within a pattern are always assigned a specific object type.
% %%
% In the case of \code/pat/, because \code/x/ occurs as second argument of a \code/ValueSnak/, KIF knows it must be a value, and thus not a statement or snak (see Figure~\ref{fig:grammar}).
% %%
% What happens is that, at line~11, when the value-snak is constructed, variable \code/x/ is coerced into a homonymous variable of type ``value''.
% %%
% KIF ensures that homonymous variables always have compatible types within a same pattern.

Before detailing how the \code{match()} call in line~11 works (see Section~\ref{sub:the-match-call}), let us make a quick detour to show two other features of KIF.%, namely, its vocabulary module and its support for constraints.

\enlargethispage*{4pt}

\subsubsection{The vocabulary module}

When writing data model or pattern objects, we can use KIF's vocabulary module to make the code less verbose.
For example, we can rewrite the previous pattern \code/pat/ (lines~9--10) more concisely as follows:
\pyconfile[firstline=12,lastline=13,linenos=true]{code/code.pycon}%
At line~12, we import KIF's Wikidata vocabulary module \code/wd/.
At line~13, we use \code{wd.Q()} and \code{wd.P()} to construct the item ``benzene'' (Q2270) and property ``LD50'' (P2240) without having to write their URLs.
But we can do better:
\pyconfile[firstline=14,lastline=14,linenos=true]{code/code.pycon}%
This constructs the same pattern by applying property \code{wd.median_lethal_dose}, which is predeclared in the \code/wd/ module, as if it were a Python function to arguments \code{wd.Q(2270)} and \code/x/.
That the three versions construct exactly the same statement pattern object can be checked by a simple equality test:
\pyconfile[firstline=15,lastline=16,linenos=true]{code/code.pycon}%

\subsubsection{Constraints}

Suppose we want to restrict the statements that match the previous pattern to those with a value in the range 4000--7000 mg/kg.
We can do that by using method \code/where()/ to constraint the pattern as follows:
\pyconfile[firstline=17,lastline=19,linenos=true]{code/code.pycon}%
Method \code{where()} takes a boolean expression of variables.
Here the resulting pattern, \code/pat3/ (lines~17--19), is a new pattern equal to \code/pat/ with the added constraint that the matched statement's value must be a quantity in mg/kg with amount~\code/x/ such that $4000\le\text{\code/x/}\le7000$, and with any lower- and upper-bound values.

%%
% TODO: Equal to -> extends.
%%

This ends our detour.  We can now get back to the \code{match()} method.

\subsection{The \texttt{match()} Method}%
\label{sub:the-match-call}

Method \code{match()} is the workhorse of the store API\@.
It must be implemented by all store types and is the basis of the implementation of most other store API methods, including \code{filter()}.

The \code{match()} method takes a pattern $p$ as argument and returns a \emph{match} object which when iterated generates all instances of $p$ found in the store.
%%
% Although most of the time $p$ is a statement pattern, \code{match()} accepts any kind of pattern.
% %%
% For instance, if we give it the pattern $\text{\code{Quantity(8, Variable('u'))}}$ it matches all quantities with amount 8 and unit $u$ (any) in the store, independently of where such a quantity occurs (statement, qualifier, or reference).
%%
%%
The actual implementation of \code{match()} varies from store to store.
But the general idea can be described as follows.
First, the store compiles pattern $p$ into a query~$q$, written in query language of knowledge source.
%%
% (This compilation should take into account what pattern $p$ is trying to accomplish, e.g., if $p$ is a statement pattern, then $q$ should be a query that looks for things that are to be interpreted as Wikidata-like statements in the source.)
%%
Then it evaluates $q$ over the source, producing a result set $R$ such that each result $b$ in~$R$ is a binding of variables in~$q$.
Finally, for each variable-value pair $(x,v)$ in~$b$, the store replaces the variable corresponding to $x$ in $p$ by the value $v$, generating a new match.

% To make matters more concrete, consider process of pattern compilation and evaluation used by the SPARQL store, illustrated in Figure~\ref{fig:pattern-evaluation}.

To make matters more concrete, let $p$ be pattern \code/pat3/ defined at the end of Section~\ref{sub:patterns} (lines~17--19).
Figure~\ref{fig:pattern-evaluation} shows the steps taken by a SPARQL store evaluate the call \code/match(!$p$!)/.

In step~(1), the SPARQL store instantiates a new SPARQL compiler.
Since no SPARQL mapping was given to the compiler, it assumes a default mapping targeting the RDF encoding of Wikidata.

In steps~(2) and~(3), the store uses the compiler to compile pattern $p$ into a SPARQL query~$q$ and a substitution~$\theta$.
Note that compilation is compositional, i.e., the compilation of~$p$ is defined in terms of the compilation of its subpatterns.
The substitution $\theta$ is a mapping from subpatterns of~$p$ into variables of~$q$.
For instance, the $\theta$ of Figure~\ref{fig:pattern-evaluation} specifies that variable \code/?x/ of query~$q$ corresponds to variable \code/x/ of pattern~$p$.

In steps~(4) and~(5), the SPARQL store sends query~$q$ to the source's SPARQL endpoint and receives as a result the SPARQL result $R$.
A SPARQL result is essentially a set of bindings of the variables selected by the query.
Figure~\ref{fig:pattern-evaluation} shows that $R$ contains at least two bindings for variable \code/?x/, namely, 4700 and 6400.

\makeatletter
\newcommand\notsotiny{\@setfontsize\notsotiny\@vipt\@viipt}
\makeatother
\begin{figure}[h]
  %\vskip-\baselineskip%
  \centering
  \begin{tikzpicture}[
    font=\rmfamily\footnotesize,
    node distance=2em,
    label/.style={font=\scriptsize},
    box/.style={draw,text width=6em,align=center}]
    \node[box](Store){\strut SPARQL\\\strut Store};
    \begin{scope}[node distance=9em]
      \node[box,below=of Store](Compiler){\strut SPARQL\\\strut Compiler};
    \end{scope}
    \begin{scope}[node distance=21em]
      \node[box,right=of Store](Endpoint){\strut SPARQL\\\strut Endpoint};
    \end{scope}
    \coordinate(C1) at ($(Store.north)+(-1,1.8em)$);
    \coordinate(C2) at (C1|-Store.south);
    \coordinate(C6) at ($(Store.north)+(1,1.8em)$);
    \coordinate(C3) at (C6|-Store.south);
    \coordinate(C4) at ($(Store.east)+(0,-.4)$);
    \coordinate(C5) at ($(Store.east)+(0,.4)$);
    \draw[->](C1) to node[label,anchor=east]{\strut(1)~\code/match(!$p$!)/} (C1|-Store.north);
    \draw[->](C2) to node[label,anchor=east]{\strut(2)~$p$} (C2|-Compiler.north);
    \draw[<-](C3) to node[label,anchor=west]{\strut(3)~$q,\theta$}
             (C3|-Compiler.north);
    \draw[->](C4) to node[label,below]{\strut(4)~$q$} (C4-|Endpoint.west);
    \draw[<-](C5) to node[label,above]{
      \strut(5)~$R=\{(\text{\code/?x/},4700),(\text{\code/?x/},6400),\ldots\}$}
             (C5-|Endpoint.west);
    \draw[<-](C6) to node[label,anchor=west]{
      \strut(6)~$
      \theta[\text{\code/?x/}\coloneq{4700}](p),\
      \theta[\text{\code/?x/}\coloneqq{6400}](p),\ \ldots$}
    (C6|-Store.north);
    \node[overlay,below=of Endpoint.east,anchor=north east,yshift=1em,
          font=\notsotiny,text width=26em]{%
      where:\\
      \begin{tabular}{rl}
        $p=$&\code/wd.median_lethal_dose(wd.Q(2270),/\\
            &\code/Quantity(x, wd.milligram_per_kilogram).where(x.ge(4000) & x.le(7000)))/\\
        $\theta=$&$\{(\text{\code/Variable('x')/},\text{\code/?x/})\}$\\
        $q=$&\code/SELECT * WHERE {/\\
            &\code/wd:P2240 wikibase:claim ?_v4 .             # ?_v4 := p:P2240/\\
            &\code/wd:P2240 wikibase:statementProperty ?_v0 . # ?_v0 := ps:P2240/\\
            &\code/wd:P2240 wikibase:statementValue ?_v3 .    # ?_v3 := psv:P2240/\\
            &\code/wd:Q2270 ?_v4 ?_v1 .                       # ?_v1 := wds:_/\\
            &\code/?_v1 ?_v3 ?_v2 .                           # ?_v2 := wdv:_/\\
            &\code/?_v1 ?_v0 ?x ./\\
            &\code/?_v2 rdf:type wikibase:QuantityValue ./\\
            &\code/?_v2 wikibase:quantityAmount ?x ./\\
            &\code/?_v2 wikibase:quantityUnit wd:Q21091747 ./\\
            &\code/OPTIONAL { ?_v2 wikibase:quantityLowerBound ?_v5 . }/\\
            &\code/OPTIONAL { ?_v2 wikibase:quantityUpperBound ?_v6 . }/\\
            &\code/FILTER (?x >= 4700 && ?x <= 7000) }/
      \end{tabular}};
  \end{tikzpicture}
  \caption{Evaluation of \code{match(!$p$!)} over a SPARQL store.}%
  \label{fig:pattern-evaluation}%
  \vskip-\baselineskip%
\end{figure}


%%% Local Variables:
%%% mode: latex
%%% TeX-engine: xetex
%%% TeX-master: "../main"
%%% eval: (visual-line-mode 1)
%%% End:

Finally, in step~(6), for each binding~$b$ in $R$, the SPARQL store replaces the SPARQL variables in $\theta$ by their values in $b$ and applies the resulting substitution to the original pattern $p$ to obtain a new match.
For example, the first match shown in step~(6) of Figure~\ref{fig:pattern-evaluation} is obtained by computing $\theta[\text{\code/?x/}\coloneq{4700}](p)$, i.e., replacing \code/?x/ by 4700 in~$\theta$ and applying the resulting substitution to~$p$.
The result in this case is statement~\eqref{eq:stmt1} of Section~\ref{sec:background}.

\subsection{SPARQL Mapping for PubChem}%
\label{sub:sparql-mapping-for-pubchem}

The SPARQL store can be used to read statements from any SPARQL endpoint, provided it is supplied with an appropriate mapping.
One such mapping already included in KIF is for the RDF distribution of PubChem~\cite{Fu-G-2015,Kim-S-2023}.
Here is how we create a SPARQL store pointing to a local installation of PubChem's RDF\@:
\pyconfile[firstline=20,lastline=22,linenos=true]{code/code.pycon}%

We won't go into detail here, but object \code/PubChemMapping()/ (line~22) tells the SPARQL store (actually, the SPARQL compiler) how to translate KIF patterns into graph patterns over PubChem's RDF graph.
The resulting store, \code/PubChem/ (line~21), behaves as an ordinary SPARQL store.
We can use it, for example, to obtain the mass (P2067) of benzene from PubChem:
\pyconfile[firstline=23,lastline=28,linenos=true]{code/code.pycon}%

There are a couple of things to note here.
First, the PubChem mapping provided by KIF adopts the Wikidata vocabulary whenever possible.
For instance, it maps property ``mass'' (P2067) to the appropriate property in PubChem.
The mapping also uses Wikidata units, e.g., ``dalton'' (Q483261) above.
What it doesn't do is translate the ids of PubChem compounds.
This explains the non-Wikidata URL in the subject of the returned statement (line~27).

This is also the reason we didn't use the Wikidata id of benzene, \code/wd.Q(2270)/, in the subject of the \code/filter()/ call above.
Had we done that the result would be empty.
Instead, we used the value-snak \code/wd.InChIKey('UHOV!\dots!')/.
That is, we set the subject to any entity whose InChIKey (P235) is equal to ``UHOV\dots'', which happens to be the InChIKey of benzene.
InChIKey is a universal identifier for chemical compounds which is defined in both Wikidata and PubChem.

Here is the pattern corresponding to the previous \code/filter()/ (lines~23--25):
\pyconfile[firstline=30,lastline=32,linenos=true]{code/code.pycon}%
It now should be clear what is happening: \code/filter()/ is searching for statements with subject~\code/x/, property ``mass'', and value~\code/y/, such that~\code/x/ is an entity with ``InChIKey'' equal to ``UHOV\dots''.

This example illustrates the use of a statement pattern (line~32) as a boolean constraint, i.e., as an argument to the \code/where()/ call.
The constraint in this case plays the role of a \emph{fingerprint}, i.e., a test that identifies an entity indirectly, here using a universal identifier instead of a local one.
The support for this kind of fingerprinting technique is crucial for enabling meaningful queries over combinations of knowledge sources, such as those obtained via mixer stores.

\subsection{The Mixer Store}%
\label{sub:the-mixer-store}

The mixer store combines one or more child stores into a new store which behaves as their virtual union.
For example, we can combine \code/Wikidata/ (line~2) and \code/PubChem/ (lines~21--22) into a new store \code/mix/ of type ``mixer'' as follows:
\pyconfile[firstline=33,lastline=35,linenos=true]{code/code.pycon}%
From the user's point of view, \code/mix/ (line~33) is a store like any other.
Its content can be thought of as the union of the statements in \code/Wikidata/ and \code/PubChem/.

At lines 34--35, we ask \code/mix/ for statements with subject ``benzene'', and assign the first two results to variables \code/stmt1/ and \code/stmt2/, respectively.
One possibility here, for example, is that \code/mix/ returns first a statement from \code/Wikidata/ and then a statement from \code/PubChem/, say, those in lines~6--7 and~26--28.

% One possibility here, for example, is that \code/mix/ returns first from \code/Wikidata/ the statement in lines~6--7 (LD50 equals 4700 mg/kg) and then from \code/PubChem/ the statement in lines~26--28 (mass equals 78.11 dalton).

One way to determine which statement came from which child store, is to instruct them to attach extra references to their statements.
For instance, here is how we can instruct \code/Wikidata/ to attach an extra reference to statements:
\pyconfile[firstline=36,lastline=37,linenos=true]{code/code.pycon}%
Now every statement produced by the \code/Wikidata/ store will be associated to one extra reference stating that the statement's ``reference URL'' (P854) is the address of the endpoint set in \code/Wikidata/.
We won't go into details here, but the references of a statement can be obtained using the store API method \code{get_annotations()}.

To decouple statements from qualifiers, references, and rank---and avoid opaque ids---KIF introduces the notion of an annotation.
An \emph{annotation} (or \emph{annotation record}) is a triple containing qualifiers (set of snaks), references (set of reference records), and rank.

In KIF, statements are identified by their content (subject and snak) and can be associated with one or more annotation records in a store.
This deviates from the Wikidata RDF representation~\cite{Wikibase-RDF-Dump-Format}, in which statements are identified by opaque ids which carry its qualifiers, references, and rank.
The rationale of KIF's approach is to relieve users from having to deal with opaque, meaningless ids---push this work to the framework.

\subsection{Other Store Types and Methods}%
\label{sub:other-store-types-and-methods}

Besides the SPARQL store and the mixer store, KIF comes with an RDF store and a CSV store.
The RDF store reads statements from RDF files.
It is essentially a SPARQL store that uses RDFLib~\cite{RDFLib} to load RDF files and evaluate SPARQL queries over their contents.
The CSV store reads statements from CSV files.
It expects a mapping specifying how line-columns pairs are to be interpreted as statements.
Currently, the CSV store is implemented as a wrapper to the RDF store which first converts the CSV to RDF before loading it.
We have plans for a more direct, non-RDF-based implementation though.

All stores implement a common store API, containing core methods \code{filter()} and \code{match()} discussed before plus methods to get statement annotations (qualifiers, references, and ranks) and entity descriptors (labels, aliases, and descriptions).
The store API also has convenience methods for testing and counting pattern occurrences, and methods for obtaining store metadata.
For more details, see the documentation of KIF~\cite{KIF}\@.

% LocalWords:  KIF CSV WDQS firstline lastline linenos kif sparql eq stmt
% LocalWords:  snak LD KIF's wd PubChem PubChem's PubChemMapping dalton
% LocalWords:  UHOV InChIKey snaks RDFLib


%%% Local Variables:
%%% mode: latex
%%% TeX-engine: xetex
%%% TeX-master: "main"
%%% eval: (visual-line-mode 1)
%%% End:

\section{Use Case and Evaluation}%
\label{sec:use-case-and-evaluation}

The research on KIF was in part motivated by the development of a chat-bot for the domain of chemistry.
Depending on the user's question, the bot would retrieve statements about chemical compounds from various sources and present them as ``known facts''.
There were three main requirements: (i)~the retrieved statements should be comparable, i.e., they should use the same vocabulary when talking about the same things; (ii)~their origin should be traceable, i.e., statements should come with provenance information; and (iii)~it should be easy to add new sources to the system.

Figure~\ref{fig:use-case} shows the instantiation of KIF used in the chat-bot application.
A mixer store is used to combine three SPARQL stores: one pointing to a local installation of PubChem's RDF~\cite{Fu-G-2015} running on Virtuoso~\cite{Erling-O-2012}, one pointing to Wikidata's public SPARQL endpoint (WDQS), and one pointing to a local SPARQL endpoint built by Ontop over the SQL endpoint of IBM CIRCA\@.

\begin{figure}[b]
  \centering
  \begin{tikzpicture}[
    node distance=.5em,font=\rmfamily\footnotesize,
    cyl/.style={cylinder,aspect=.15,shape border rotate=90,inner sep=2pt}]
    \node[draw,minimum width=22em,minimum height=2em](mx){\strut Mixer};
    \begin{scope}[minimum width=7em, minimum height=2em,font=\scriptsize]
      \node[draw,below=of mx.south west,anchor=north west](S1){SPARQL Store};
      \node[draw,below=of mx.south,anchor=north](S2){SPARQL Store};
      \node[draw,below=of mx.south east,anchor=north east](S3){SPARQL Store};
    \end{scope}
    \begin{scope}[node distance=2em]
      \node[draw,minimum width=7em, minimum height=2em,below=of S3]
      (Ontop){\strut Ontop};
    \end{scope}
    \begin{scope}[node distance=1em,overlay,font=\scriptsize]
      \node[left=of S1,text width=3.75em,align=center](MS1){mapping\\(KIF)};
      \node[right=of Ontop,text width=3.75em,align=center]
      (MOntop){mapping\\(R2RML)};
      \draw[->](MS1) to (S1);
      \draw[->](MOntop) to (Ontop);
    \end{scope}
    \begin{scope}[font=\scriptsize,node distance=.5em]
      \node[draw,cyl,below=of Ontop,text width=3em,align=center]
      (CIRCA){IBM CIRCA};
      \node[draw,cyl,text width=3em,align=center]
      (Wikidata) at (S2|-CIRCA) {Wiki\-data};
      \node[draw,cyl,text width=3em,align=center]
      (PubChem) at (S1|-CIRCA) {Pub\-Chem};
    \end{scope}
    \draw(PubChem) to (S1);
    \draw(Wikidata) to (S2);
    \draw(CIRCA) to (Ontop);
    \draw(Ontop) to (S3);
    \coordinate(Y) at ($(Ontop)!.5!(S3)+(9,0)$);
    \coordinate(X) at ($(S1|-Y)+(-9,0)$);
    \draw[dashed](X) -- (Y);
    \begin{scope}
      \node[anchor=north west,yshift=-.25em] at (X){knowledge sources};
      \node[anchor=south east,yshift=.25em] at (Y){KIF};
    \end{scope}
  \end{tikzpicture}
  \caption{KIF instantiation integrating PubChem, Wikidata, and IBM CIRCA.}%
  \label{fig:use-case}
\end{figure}


%%% Local Variables:
%%% mode: latex
%%% TeX-engine: xetex
%%% TeX-master: "../main"
%%% eval: (visual-line-mode 1)
%%% End:

IBM CIRCA~\cite{IBM-CIRCA} is a relational database of chemical data extracted from patents and other sources.
In the chat-bot application, we used Ontop to build a Wikidata-compatible SPARQL endpoint to access parts of its schema.
Ontop~\cite{Xiao-G-2020} is an ontology-based data access tool.
It uses R2RML~\cite{Cyganiak-R-2012} mappings to translate SPARQL queries to SQL at query time.
The R2RML mappings tell Ontop how to map tables in the relational database into concepts and properties of the target ontology (in our case, Wikidata's).

\subsection{Evaluation}

For the evaluation, we used the setup shown in Figure~\ref{fig:use-case} without the IBM CIRCA part, i.e., essentially the setup shown in line~33 of Section~\ref{sub:the-mixer-store}.
Our goal was to measure the overhead of KIF when evaluating simple application-level queries over the mixer.
By overhead, we mean time spent in KIF (Python code) versus time spent in the SPARQL endpoints (outside the Python code).
By simple application-level queries, we mean statement patterns meaningful to users.

For the experiment, we generated patterns of the forms
\[
  \text{\code/!$p$!(x,!$v_1$!).where(!$q$!(x,!$v_2$!))/}
  \quad\text{and}\quad
  \text{\code/!$p$!(!$v_1$!,x).where(!$q$!(x,!$v_2$!))/},
\]
where \code/x/ is a pattern variable, $p$ and $q$ are properties, $v_1$ is a value or variable, and $v_2$ is a value.
Since we wanted matches in both Wikidata and PubChem, we restricted $p$ to the properties in PubChem whose domain or range is a chemical compound (e.g., mass (P2067), has part (P527), trading name (P6427), etc.) and $q$ to those which are compound identifiers (e.g., InChIKey (P235), canonical SMILES (P233), ChEBI ID (P683), etc.).

\begin{figure}[ht]
  \centering
  \relax{\includegraphics[
    clip,
    trim=4.1em .5em 5.9em 2.2em,%lbrt
    width=.9\textwidth]{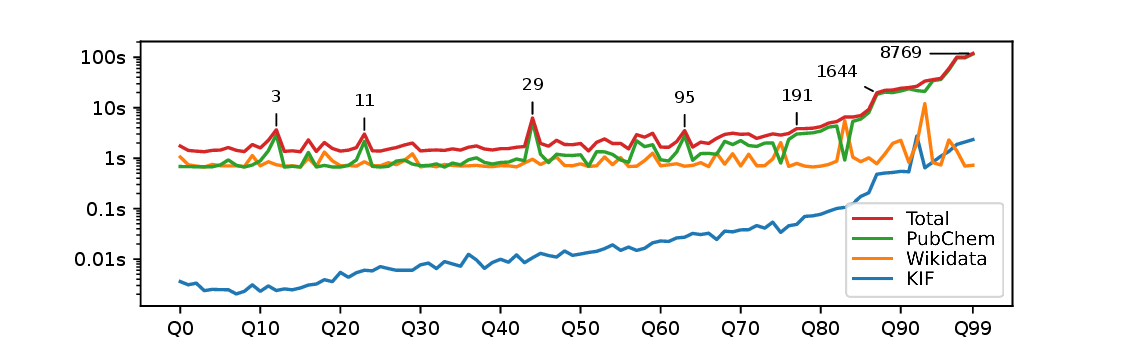}}%
  \caption{%%
    KIF overhead.
    The x-axis shows the queries sorted by number of matches.
    The numbers above the red line indicate the number of matches for a particular query.
    The y-axis shows the time in seconds (in log scale).
    On average, 2.1\% of the time was spent in KIF, 12.4\% in Wikidata, and 85.5\% in PubChem.}%
  \label{fig:plot}
\end{figure}

We evaluated each of the generated patterns over the mixer of Figure~\ref{fig:use-case} and collected those which (i) matched at least one statement in both Wikidata and PubChem, and (ii) took at least 0.75s to run in each of the endpoints.
We then sorted the patterns by number of matches and selected the last~100 patterns.
We ended up with patterns Q0--Q99, whose evaluation times are shown in Figure~\ref{fig:plot}.

For each pattern $Q_i$ in Figure~\ref{fig:plot}, the red line (Total) indicates the time taken to evaluate and consume all results of the call \code{match(!$Q_i$!)} over the mixer.
The orange and green lines (Wikidata and PubChem) indicate the time taken to send the SPARQL queries to the endpoints and receive the results.
The blue line (KIF) indicates the time spent in KIF where $\text{KIF}=\text{Total}-(\text{Wikidata}+\text{PubChem})$.

As expected, the bulk of the time (on average 97.9\%) was spent in the endpoints, in particular, PubChem.
The overhead of KIF is negligible, especially when the number of matches is smaller than the page-size configured in KIF\@.
In this experiment, we used a page of size~100, which is the default.
This means that to consume a response with 1000 matches, KIF has to perform 10 requests.
This explains the increase in the overhead of KIF after~Q70.
Nevertheless, evaluation time is dominated by the SPARQL endpoints.

% LocalWords:  KIF PubChem's WDQS Ontop RML PubChem InChIKey ChEBI


%%% Local Variables:
%%% mode: latex
%%% TeX-engine: xetex
%%% TeX-master: "main"
%%% eval: (visual-line-mode 1)
%%% End:

\section{Related Work}%
\label{sec:related-work}

There are three classes of work related to KIF\@.

%\bigskip

First, there are the ontology-based data access (OBDA) systems~\cite{Xiao-G-2018}.
These are systems such as Ontop~\cite{Xiao-G-2020} which, given a mapping between a data source and target ontology, are able to evaluate over the data source queries written in terms of the ontology.
In ODBA, the data source is a relational database, the ontology is written in a DL-Lite~\cite{Calvanese-D-2007} language (e.g., OWL~2 QL), the query language is SPARQL, and query evaluation is done under the SPARQL~1.1 entailment regime~\cite{Xiao-G-2018}.
Also, OBDA usually means virtual access: the SPARQL query is transformed into an SQL query on-the-fly and evaluated over the database.
(To be feasible, the SQL query needs to be heavily optimized, as the transformation from SPARQL often leads to a blow-up in size~\cite{Gu-Z-2022}.)

Different from OBDA systems, KIF fixes the target syntax and (possibly) the vocabulary to those of Wikidata.
Although it doesn't attempt to do any kind of reasoning on its own, its SPARQL store can be used seamlessly with any reasoning-enabled endpoint.
Also, KIF doesn't attempt to provide an interface for arbitrarily complex queries.
%%
% Instead, it provides a pattern language defined in terms of Wikidata's data model which inherits its restrictions.
%%
The \code{filter()} method, in particular, was inspired by the work on linked data fragments~\cite{Verborgh-R-2014} and TPF~\cite{Verborgh-R-2016}.
It is a lightweight filtering interface which can be used reliably by applications.

Another thing that distinguishes the KIF from OBDA systems is that the latter largely ignore the problem of query federation~\cite{Gu-Z-2022}.
(One notable exception is Squerall~\cite{Mami-M-N-2019}.)
That said, OBDA systems such as Ontop can be used to enable KIF to interface with relational databases, as illustrated in Section~\ref{sec:use-case-and-evaluation}.

%\bigskip

The second class of work related to KIF are RDF integration systems based on SPARQL query-rewriting~\cite{Correndo-G-2010,Makris-K-2012}.
These are similar to OBDA systems but target RDF\@.
The system is given an RDF source, a target ontology, a mapping between the source schema and the target ontology, and a SPARQL query written in terms of the ontology.
Its job is to translate the original SPARQL query into a new SPARQL query and evaluated it over the RDF source on-the-fly.

The problem of SPARQL-rewriting (or ontology-mediated query translation) is closely related to the ontology matching problem.
But despite the vast literature on ontology matching~\cite{Osman-I-2021}, there is little research on using the produced mappings for querying RDF sources, especially when federations of sources are considered~\cite{Cheng-S-2024}.
An early work that uses SPARQL rewriting for integrating multiple RDF sources is~\cite{Makris-K-2012}.
However, it doesn't cover the result reconciliation problem, i.e., using the mappings to reconcile the results of the queries in the terms of the target ontology.
A more recent proposal which covers query-translation and result reconciliation is~\cite{Cheng-S-2024}.
The previous remarks on the differences between KIF and OBDA systems also apply to SPARQL rewriting systems.

%\bigskip

The third class of related work are knowledge graph construction (KGC) systems.
These are systems like SPARQL-Generate~\cite{Lefrancois-M-2017} and SPARQL Anything~\cite{Asprino-L-2023}, which use SPARQL to describe the transformation of non-RDF sources into an RDF dataset.
For this purpose, SPARQL-Generate extends the syntax of SPARQL, while SPARQL Anything overrides the SERVICE clause.
Both of these systems allow users to query non-RDF resources on-the-fly.
But different from KIF, they don't attempt perform any kind of ontology-mediated mapping.
The user must use the vocabulary of each source and specify the desired transformations explicitly, for each query.

%\bigskip

One problem largely ignored by the three classes of work cited above is the representation and tracking of provenance.
In contrast, KIF tackles this problem from the start: statements carry provenance information which is preserved while they traverse the framework.

% LocalWords:  KIF OBDA Ontop ODBA DL QL TPF Squerall KGC


%%% Local Variables:
%%% mode: latex
%%% TeX-engine: xetex
%%% TeX-master: "main"
%%% eval: (visual-line-mode 1)
%%% End:

\section{Conclusion}%
\label{sec:conclusion}

\enlargethispage*{3pt}

This paper presented KIF, a framework that uses Wikidata to virtually integrate heterogeneous knowledge sources.
Having Wikidata as the integration model brings some advantages, including a flexible data model and access to a huge vocabulary covering a wide range of topics.
This is particularly important when the integration problem is open-ended: when we don't know which kind of knowledge source we might need to integrate next, or when the sources deal with completely different subjects.
In the domain of chemistry, as illustrated in Section~\ref{sec:background}, when talking about toxicity we often need to refer not only to chemicals but also to things like symptoms, diseases, and even animals---all of these are already covered by Wikidata's vocabulary.
That said, KIF is not restricted to Wikidata's vocabulary.
If needed, the stores can extend their vocabulary with new entities and properties (cf.~the use of PubChem URLs as item ids by the PubChem RDF mapping in Section~\ref{sub:sparql-mapping-for-pubchem}).

% Regarding the architecture of KIF, we tried to conciliate ease of use with flexibility and adherence to standards (SPARQL and RDF).
% %%
% The latter helps to promote the reuse of existing solutions.
% %%
% For example, we currently use Ontop~\cite{Xiao-G-2020} and Comunica~\cite{Taelman-R-2018} in the KIF Middleware, respectively, to construct SPARQL endpoints over SQL endpoints and to aggregate multiple SPARQL endpoints into a single endpoint.
% %%
% We are also studying the integration of other similar solutions: HeFQUIN~\cite{Cheng-S-2024}, as a SPARQL-rewriting-enabled alternative to Comunica, and Squerall~\cite{Mami-M-N-2019}, as a query federation-enabled alternative to Ontop.

Moving to the implementation of KIF, we are currently adding support for parallel requests to the library.
This will speed up not only the mixer but also any store that needs to split large queries into multiple requests.
As we mentioned in Section~\ref{sub:other-store-types-and-methods}, we are also working on an alternative, non-RDF-based implementation of the CSV store.
A more important change is the addition of a mutable store API, which will allow users write onto stores.
The idea is to use SPARQL update queries in the case of the SPARQL store.

Finally, on the theoretical side we are working on the formalization of the semantics of KIF patterns, inspired by the work on WShEx~\cite{Gayo-J-E-L-2022}.

% LocalWords:  KIF PubChem CSV WShEx


%%% Local Variables:
%%% mode: latex
%%% TeX-engine: xetex
%%% TeX-master: "main"
%%% eval: (visual-line-mode 1)
%%% End:

\bibliographystyle{splncs04}
\bibliography{bib}

\end{document}